%% file: main.tex
\documentclass{article}

\usepackage{hyperref}       
\usepackage{url}            
\usepackage{graphicx}
\usepackage{subfigure}
\usepackage{subfiles}
\usepackage{tabularx}
\usepackage{etoolbox}
\usepackage[dvipsnames]{xcolor}
\usepackage{bbm}
\usepackage{multirow}
\usepackage{enumitem}
\usepackage{amssymb}
\usepackage{amsmath}
\usepackage{amsthm}
\usepackage{xspace}
\usepackage{float}
\usepackage{booktabs}       
\usepackage{amsfonts}       
\usepackage{nicefrac}       
\usepackage{microtype}      
\usepackage{color, soul}

\usepackage{setspace}
\usepackage{algorithm}
\usepackage{algpseudocode}

\usepackage{algorithm}
\usepackage{algpseudocode}
\usepackage{math_commands}

\newtoggle{arxiv}
\toggletrue{arxiv}

\newcommand{\yell}[1]{{\color{black}#1}}

\iftoggle{arxiv}{
    \usepackage[numbers,sort]{natbib}
    \setlength{\textwidth}{6.5in}
    \setlength{\textheight}{9in}
    \setlength{\oddsidemargin}{0in}
    \setlength{\evensidemargin}{0in}
    \setlength{\topmargin}{-0.5in}
    \newlength{\defbaselineskip}
    \setlength{\defbaselineskip}{\baselineskip}
    \setlength{\marginparwidth}{0.8in}
}

\newcolumntype{Y}{>{\hsize=.7\hsize}X}
\newcolumntype{Z}{>{\hsize=1.3\hsize}X}

\makeatletter
\def\@copyrightspace{\relax}
\makeatother

\makeatletter
\def\@myauthornotes{}
\def\myauthornote#1{%
  \if@ACM@anonymous\else
    \g@addto@macro\addresses{}%
    \g@addto@macro\@myauthornotes{%
      \stepcounter{footnote}\footnotetext{#1}}%
  \fi}
\makeatother

\iftoggle{arxiv}{
    \title{Metadata Shaping: Natural Language Annotations for the Tail}
    \usepackage{authblk}
    \author[]{Simran Arora}
    \author[]{Sen Wu}
    \author[]{Enci Liu}
    \author[]{Christopher R{\'e}}
    \affil[]{Stanford University}
    \affil[ ]{\texttt{\{simran,senwu,jesslec,chrismre\}@cs.stanford.edu}}
}

\begin{document}
\maketitle

\begin{abstract}
\label{sec:abstract}
\subfile{Sections/abstract}
\end{abstract}

\section{Introduction}
\input{Sections/intro.tex}

\section{Method}
\input{Sections/method.tex}

\section{Experiments}
\input{Sections/experiments.tex}

\section{Analysis}
\input{Sections/principles.tex}

\section{Discussion}
\input{Sections/discussion.tex}

\section{Related Work}
\input{Sections/related_work.tex}

\section{Conclusion}
\input{Sections/conclusion.tex}

\subsubsection*{Acknowledgments}
We thank Mayee Chen, Neel Guha, Megan Leszczynski, and Laurel Orr for their helpful feedback. We gratefully acknowledge the support of NIH under No. U54EB020405 (Mobilize), NSF under Nos. CCF1763315 (Beyond Sparsity), CCF1563078 (Volume to Velocity), and 1937301 (RTML); ONR under No. N000141712266 (Unifying Weak Supervision); ONR N00014-20-1-2480: Understanding and Applying Non-Euclidean Geometry in Machine Learning; N000142012275 (NEPTUNE); the Moore Foundation, NXP, Xilinx, LETI-CEA, Intel, IBM, Microsoft, NEC, Toshiba, TSMC, ARM, Hitachi, BASF, Accenture, Ericsson, Qualcomm, Analog Devices, the Okawa Foundation, American Family Insurance, Google Cloud, Salesforce, Total, the HAI-AWS Cloud Credits for Research program, the Stanford Data Science Initiative (SDSI), the Stanford Graduate Fellowship in Science and
Engineering, and members of the Stanford DAWN project: Facebook, Google, and VMWare. The Mobilize Center is a Biomedical Technology Resource Center, funded by the NIH National Institute of Biomedical Imaging and Bioengineering through Grant P41EB027060. The U.S. Government is authorized to reproduce and distribute reprints for Governmental purposes notwithstanding any copyright notation thereon. Any opinions, findings, and conclusions or recommendations expressed in this material are those of the authors and do not necessarily reflect the views, policies, or endorsements, either expressed or implied, of NIH, ONR, or the U.S. Government.

\bibliographystyle{plainnat}
\bibliography{references}

\appendix
\label{sec:appendix}
\input{Sections/appendix}

\end{document}

%% file: Sections/abstract.tex
Language models (LMs) have made remarkable progress, but still struggle to generalize beyond the training data to rare linguistic patterns. Since rare entities and facts are prevalent in the queries users submit to popular applications such as search and personal assistant systems, improving the ability of LMs to reliably capture knowledge over rare entities is a pressing challenge studied in significant prior work. Noticing that existing approaches primarily modify the LM architecture or introduce auxiliary objectives to inject useful entity knowledge, we ask to  what extent  we  could  match the  quality  of  these  architectures  using a base LM architecture,  and only changing the data? We propose \textit{metadata shaping}, a method in which readily available metadata, such as entity descriptions and categorical tags, are appended to examples based on information theoretic metrics. Intuitively, if metadata corresponding to popular entities overlap with metadata for rare entities, the LM may be able to better reason about the rare entities using patterns learned from similar popular entities. On standard entity-rich tasks (TACRED, FewRel, OpenEntity), with no changes to the LM whatsoever, metadata shaping exceeds the BERT-baseline by up to \yell{5.3 F1} points, and achieves or competes with state-of-the-art results. We further show the improvements are up to \yell{10x} larger on examples containing tail versus popular entities.

%% file: Sections/intro.tex
While recent language models (LMs) are remarkable at learning patterns seen frequently during training, they still suffer from performance degradation  over rare \textit{tail} patterns. 
In this work, we propose metadata shaping, a method to address the tail challenge by encoding properties shared between popular and tail examples \textit{in the data itself}. 

\begin{figure}[t!]
    \centering
    \includegraphics[width=\linewidth]{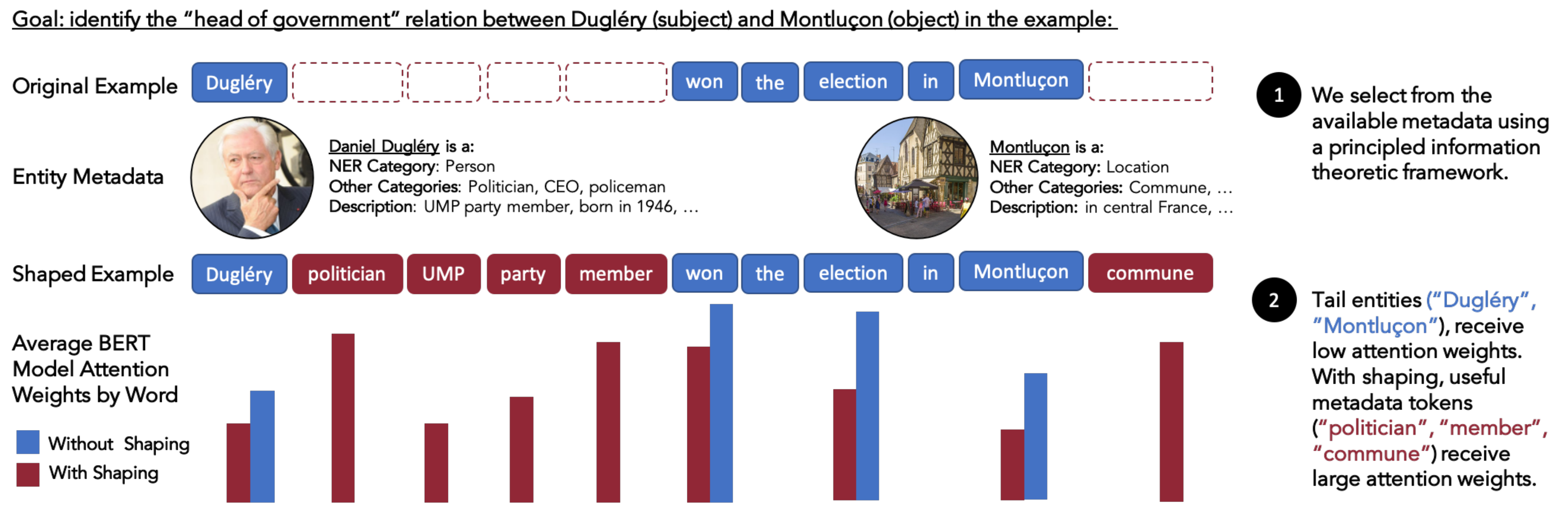}
    \caption[width=0.9\linewidth]{Metadata shaping inserts metadata (e.g., entity categories and descriptions) into the data at train and test time. The FewRel benchmark involves identifying the relation between a subject and object string. The above subject and object are unseen in the FewRel training data and the tuned base LM places low attention weights on those tokens. A base LM trained with shaped data has high attention weights for useful metadata tokens such as ``politician''. Weights are shown for words which are not stop-words, punctuation, or special-tokens.}
    \label{fig:shaping_main}
\end{figure}

Tail degradation is particularly apparent in \textit{entity-rich tasks}, since users care about very different entities and facts when using popular entity-centric applications such as search and personal assistants \citep{bernstein2012direct, perner2020ebert, bootleg}.
Given the importance of entity-rich applications, considerable effort has been devoted to methods for capturing entity knowledge more reliably. Many proposed approaches proceeding by changing the model architecture.

While early LMs relied heavily on feature engineering \citep{lafferty2001crf, zhang2011ftreng}, recent neural models could learn useful features during training \citep{collobert2011nlpscratch}, and shifted the focus to model engineering. As such, given data annotated with entity tags, recent methods for reasoning about entities typically enrich \textit{base LMs} (e.g., \citet{devlin2018bert}) by modifying the architecture or introducing auxiliary objectives to better learn entity properties \citep{zhang2019ernie, peters2019knowledge, logan2019kglm, yamada2020luke, wang2020kepler, xiong20wklm, su2021cokebert}, or by combining multiple learned modules, each specialized to handle fine-grained reasoning patterns or subsets of the data distribution \citep{chen2019slice, wang2020kadapter}. 

These \textit{knowledge-aware LMs} have led to impressive gains compared to base LMs on entity-rich tasks, however they raise a few challenges: additional pretraining is expensive, deploying and maintaining multiple specialized LMs can be memory and time-intensive, each LM needs to be optimized for efficient use (e.g., on-device), and the methods require training to adapt to changing facts or new entities. Further, these LMs learn about an entity from its individual occurrences during training, rather than explicitly encoding that patterns learned for one entity may be useful for reasoning about other \textit{similar} entities. Implicitly learning subpopulations may be data-inefficient as the Wikidata knowledge base alone holds \yell{$\sim 100$M} entities and unfortunately, \yell{89\%} of the Wikidata entities do not even appear in Wikipedia, a popular source of unstructured training data for the LMs. \footnote{\citet{bootleg} finds that a BERT based model needs to see an entity in on the order of 100 samples to achieve 60 F1 points when disambiguating the entity in Wikipedia text.} 

Thus, we asked the question: to what extent could we match the quality of these architectures using a base LM architecture, and only changing the data? We propose some simple modifications to the data at train and test time, which we call metadata shaping, and find the method is surprisingly quite effective.
Given unstructured text, there are several readily available tools for generating annotations at scale \citep{manning2014cornlp, honnibal2020spacy}, and knowledge bases contain entity metadata including categorical tags (e.g., Barack Obama is a ``politician'') and descriptions (e.g., Barack Obama ``enjoys playing basketball''). Metadata shaping entails explicitly inserting retrieved metadata in the example as in Figure \ref{fig:shaping_main} and inputting the resulting \textit{shaped example} to the LM. We find focusing on the data to be a compelling way to improve tail generalization. Our contributions are:

\paragraph{Simple and Effective Method} We propose metadata shaping and demonstrate its effectiveness on entity-rich benchmarks. Metadata shaping, with simply an \textit{off-the-shelf} base LM, exceeds the base LM trained on unshaped data by up to \yell{5.3 F1} points and competes with state-of-the-art methods, which do modify the LM. Metadata shaping thus enables re-using well-studied and optimized base LMs (e.g., \citet{sanh2020distilbert}).

\paragraph{Tail Generalization} We show that the performance improvements are concentrated on the tail --- the observed improvement from shaping for examples containing tail entities is up to 10x larger than on the slice of examples containing popular entities across the evaluated tasks.\footnote{Tail entities are those seen $< 10$ times during training, consistent with \citep{goel2021robustnessgym, bootleg}.} 
Metadata shaping simplifies the task to learning properties of \textit{groups} of entities. Metadata establish ``subpopulations'', groups of entities sharing similar properties, in the entity distribution \citep{zhu2014objecttail, cui2019classbalancetail}. 
For example on the FewRel benchmark \citep{han2018fewrel}, ``Daniel Dugléry'' (a French politician) appears \yell{0 times}, but ``politician'' entities in general appear \yell{$> 700$} times in the task training data. Intuitively, performance on a rare entity should improve if the LM has the explicit information that it is similar to an entity observed during training. This is data-efficient compared to learning properties of entities from individual occurrences.
    
\paragraph{Explainability} Prior knowledge-aware LMs use metadata \citep{peters2019knowledge, Alt2020TACREDRA}, but do not explain when and why different metadata help. There is a rich set of tools for reasoning about data distributions, in contrast to neural architectures. Inspired by classic feature selection techniques \citep{guyon2003ftrselection}, we conceptually explain the effect of metadata on generalization error.

We hope this work motivates further research on addressing the tail challenge through the data, alongside the model.

%% file: Sections/method.tex
This section introduces metadata shaping, including the set up and conceptual framework. 

\subsection{Objective} The goal of metadata shaping is to improve tail performance, using properties shared by popular and rare examples. For instance, this work investigates improving tail performance in entity-rich tasks using entity metadata (e.g., ``politician'') which are readily available for tail (e.g., ``Daniel Dugléry'') and popular \textit{head} (e.g., ``Barack Obama'') entities. We define tail entities as those seen $< 10$ times during training, consistent with \citep{goel2021robustnessgym, bootleg}.

Metadata are sourced by using off-the-shelf models (e.g., for entity linking, named entity recognition (NER), part-of-speech (POS)) \citep{manning2014cornlp, honnibal2020spacy}, heuristic rules, and metadata in  knowledge bases (KBs) (e.g., Wikidata, Wordnet \citep{miller1995wordnet}, and domain-specific KBs such as UMLS \citep{bodenreider2004umls}) to tag unstructured text, making it is easy to annotate at scale. We consider an unrestricted set of metadata, rather than a predefined schema.

\subsection{Set up}
Let $x \in \mathcal{X}$ and $y \in \mathcal{Y}$, and consider the labeled classification dataset $\pmb{D} = \{(x_i, y_i)\}_{i=1}^{n}$ of $n$ examples. Let $m \in \mathcal{M}$ denote a metadata token and let $\pmb{M(x_i)}$ be the set of metadata tokens collected for example $x_i$. A shaping function $f_s : \mathcal{X} \rightarrow \mathcal{X}_s$ accepts an original example $x_i \in \mathcal{X}$ and produces a shaped example $s_i \in \mathcal{X}_s$ by inserting a subset of $\pmb{M(x_i)}$  (See Figure \ref{fig:shaping_main}). The downstream classification model $\hat{p}_\phi$ is learned from resulting shaped data to infer $y_i$ from the shaped example $s_i$.

In this work, we use the following representative metadata shaping functions for all tasks. 
These functions range from applying signals to groups of examples to individual examples:
\begin{itemize}
    \item \textbf{Structure tokens} Structure tokens indicate the spans of interest in the task. For example, inserting the special token $[ENTITY]$ surrounding the key entities in an entity-rich task. This is a common strategy, which can be viewed as an instance of metadata shaping.
    \item \textbf{Categorical tokens} Categorical tokens create subpopulations of entities (e.g., entities like \emph{Dugléry} fall in the coarse grained \textit{person} category, or finer grained category of \textit{politician} entities). Off-the-shelf models (e.g., for NER and POS) provide coarse grained categories, and knowledge bases contain fine-grained categories (e.g., entity types and relations). Category tokens are  consistent and frequent compared to words the original example.
    \item \textbf {Descriptive tokens} Descriptions give cues for rare entities and alternate views of popular entities. E.g., \emph{Dugléry} is ``a French politician and the son of a baker''. In contrast to categories, descriptions are highly variable and likely to be unique for across entities.
\end{itemize}

\subsection{Conceptual framework}

Next we want to understand why inserting $m \in \pmb{M(x_i)}$ for $x_i \in \pmb{D}$ can improve tail performance. We measure the generalization error of the classification model $\hat{p}_{\phi}$ using the cross-entropy loss:
 
{\small\begin{align}
    \mathcal{L}_{\textup{cls}} = \mathbb{E}_{(x, y)} \big[ -\log(\hat{p}_{\phi}(y|x))\big].
\end{align}}%

Let $\Pr(y | x_i)$ be the true probability of class $y \in Y$ given $x_i$. Example $x_i$ is composed of a set of linguistic patterns $\pmb{K_i}$ (e.g., n-grams). We make the assumption that a pattern $k \in \pmb{K_i}$ is a useful signal if it informs $\Pr(y | x_i)$. We thus parametrize the true distribution $\Pr(y|x_i)$ using the principle of maximum entropy \citep{berger1996maxent}:

{\small\begin{align}
    \Pr(y|x_i) = \frac{1}{Z(x_i)} \exp(\sum_{k \in \pmb{K_i}} \lambda_{k} \Pr(y|k)).
\end{align}}%
where $\lambda_k$ represents learned parameters weighing the contributions of patterns (or events) $k$ and $Z(x_i)$ is a partition function ensuring $\Pr(y|x_i)$ represents a probability distribution. Therefore when evaluating $\hat{p}_\phi$, achieving zero cross-entropy loss between the true probability $\Pr(y|k)$ and the estimated probability $\hat{p_\phi}(y|k)$, for all $k$, implies zero generalization error overall.


\paragraph{Unseen Patterns} Our insight is that for a pattern $k$ that is unseen during training, which is common in entity-rich tasks,\footnote{For example, on the FewRel benchmark used in this work, \yell{90.7\%}, \yell{59.7\%} of test examples have a subject, object span which are unseen as the subject, object span during training.}, the class and pattern are independent ($y \perp k$) under the model's predicted distribution $\hat{p}_{\phi}$, so $\hat{p}_{\phi}(y | k) = \hat{p}_{\phi}(y)$. With the assumption of a well-calibrated model and not considering priors from the base LM pretraining stage,\footnote{For this downstream analysis, we ignore occurrences in the pretraining corpus and learned similarities between unseen $k$ and seen $k'$. Future work can incorporate those priors to refine the slice of unseen entities.} this probability is
$\hat{p}_{\phi}(y) =  \frac{1}{|\mathcal{Y}|}$ for $y \in \mathcal{Y}$. 



Plugging in $\hat{p}_{\phi}(y) =  \frac{1}{|\mathcal{Y}|}$, the cross-entropy loss between $\Pr(y|k)$ and $\hat{p}_{\phi}(y|k)$ is $\Pr(k)\log |Y|$. Our idea is to effectively replace $k$ with another (or multiple) \textit{shaped} pattern $k'$, which has non-uniform $\hat{p}_{\phi}(y|k')$ and a lower cross-entropy loss with respect to $\Pr(y|k')$, as discussed next.

\paragraph{Inserting Metadata} Consider the shaped example, $s_i=f_s(x_i)$, which contains new tokens from $\pmb{M(x_i)}$, and thus contains a new set of linguistic patterns $\pmb{K_i^s}$. Let $k_m \in \pmb{K_i^s}$ be a pattern containing some $m \in \pmb{M(x_i)}$. For the rare pattern (e.g., rare entity) $k$, if an associated pattern $k_m$ (e.g., a metadata token for the rare entity) occurs non-uniformly across classes during training, then the cross-entropy loss between $\hat{p}_{\phi}(y|k_m)$ and $\Pr(y|k_m)$ is lower than the cross-entropy loss between $\hat{p}_{\phi}(y|k)$ and $\Pr(y|k)$.
If $k_m$ provides a high quality signal, shifting $\hat{p}_{\phi}(y|x_i)$ in the correct direction, performance of $\hat{p}_\phi$ should improve. 

We can use the conditional entropy $\hat{H}(\mathcal{Y}|k)$ to measure the non-uniformity of $k_m$ across classes. When $k$ is unseen and $\hat{p}_{\phi}(y|k) = \hat{p}_{\phi}(y, k)= \hat{p}_{\phi}(y) =  \frac{1}{|\mathcal{Y}|}$ (uniform), so the conditional entropy is maximized: 

{\small\begin{align}
    \hat{H}(\mathcal{Y}|k) = - \sum_{y \in \mathcal{Y}} \hat{p}_{\phi}(y, k) \log \hat{p}_{\phi}(y|k) = \log(|\mathcal{Y}|).
\end{align}}

For non-uniform $\hat{p}_{\phi}(y|k_m)$, this conditional entropy decreases. Thus our application of using metadata for the tail connects back to classical feature selection methods \citep{guyon2003ftrselection}, we seek the metadata providing the largest information gain. Next we discuss the practical considerations for selecting metadata.

\paragraph{Metadata Shaping Selection} Entities are associated with large amounts of metadata $\pmb{M(x_i)}$ --- categories can range from coarse-grained (e.g., ``person'') to fine-grained (e.g., ``politician'' or ``US president'') and there are intuitively many ways to describe people, organizations, sports teams, and other entities. Since certain metadata may not be helpful for a task, and  popular base LMs do not scale very well to long dependencies \citep{tay2020longrangearena, pascanu2013rnnlong}, it is important to understand \textit{which} metadata to use for shaping.  

We want to select $k_m$ with non-uniform $\hat{p}_{\phi}(y|k_m)$ across $y \in \mathcal{Y}$, or in other words lower $\hat{H}(\mathcal{Y}|k_m)$. By definition, the conditional probability $\Pr(y|k_m)$ is:

{\small\begin{align}
    \Pr(y | k_m) = 2^{\pmi(y, k_m)}\Pr(y),
\end{align}}%
where we recall that the pointwise mutual information $\pmi(y, k_m)$ is defined as $\log\big(\frac{\Pr(y, k_m)}{\Pr(y) \Pr(k_m)}\big)$. The $\pmi$ compares the probability of observing $y$ and $k_m$ together (the joint probability) with the probabilities of observing $y$ and $k_m$ independently. Class-discriminative metadata reduce $\hat{H}(\mathcal{Y}|k)$.


Directly computing the resulting conditional probabilities after incorporating metadata in $\pmb{D}$ is challenging since the computation requires considering all linguistic patterns across examples, generated by including $m$. Instead we use simplistic proxies to estimate the information gain. 
In Algorithm \ref{CHalgorithm}, we focus on the subset of $\pmb{K_i^s}$ containing individual metadata tokens $m$, and computes the entropy over $\hat{p}_{\phi}(y|m)$ for $y \in \mathcal{Y}$. Simple extensions to Algorithm \ref{CHalgorithm}, at the cost of additional computation, consider a broader set of $k_m$ (e.g., $n$-grams containing $m$ for $n > 1$), or \textit{iteratively} select tokens by considering the correlations in the information gain between different metadata.


\begin{algorithm}[t!]
\caption{Metadata Token Selection}
\label{CHalgorithm}
\begin{algorithmic}[1]
\State \textbf{Precompute Train Statistics}
\State \textbf{Input}: training data $\pmb{D}_{train}$, metadata $M$
\For{each category $m \in M$ over $\pmb{D}_{train}$}
\State Compute $\pmi(y,m)$ for $y \in \mathcal{Y}$.
\EndFor
\For{each class $y \in \mathcal{Y}$ over $D_{train}$}
\State Compute frequency $f_y$.
\EndFor
\State
\State \textbf{Select Metadata for Sentence}
\State \textbf{Input}: $x_i$ from $\pmb{D}_{train}$ and $\pmb{D}_{test}$, integer $n$.
\State Collect metadata $\pmb{M(x_i)}$ for $x_i$.
\For{$m \in \pmb{M(x_i)}$}
\State Compute $r_{y} = 2^{\pmi(m, y)}f_y$ for $y \in \mathcal{Y}$.
\State Normalize $r_{y}$ values to sum to $1$.
\State Compute entropy $H_m$ over $r_{y}$ for $y \in \mathcal{Y}$.
\EndFor
\State Rank $m \in \pmb{M(x_i)}$ by $H_m$.
\State \textbf{Return} the $\min(n, |\pmb{M(x_i)}|)$ tokens with lowest $H_m$.
\end{algorithmic}
\end{algorithm}

\paragraph{Metadata Shaping Noise} Feature noising \citep{wang2013ftrnoising} is effective for structured prediction tasks to prevent overfitting. While regularization is typically applied directly to model parameters, \citet{xie2017blanknoising, dao2019kernel} regularize through the data. Since sources of metadata are not guaranteed to be accurate, and all entities may not be correctly linked or consistently categorized in the KB,  
we hypothesize that using metadata with diverse word choice and order (e.g., entity descriptions) and blank noising (e.g., by masking metadata tokens), can help reduce overfitting, and investigate this empirically in Section 4.

%% file: Sections/experiments.tex
In this section, we demonstrate that metadata shaping is general and effective, and in Section 4, we provide a detailed analysis of the framework. We apply the method to diverse entity-rich tasks and compare to current state-of-the art methods.

\subsection{Datasets}

We evaluate on standard entity-typing and relation extraction benchmarks used by baseline methods.

\paragraph{Entity Typing} The entity typing task involves predicting the the applicable \textit{types} for a given substring in the input example, from set of output types. We use the OpenEntity \citep{choi2018openentity}, which
consists of a set of nine output types and is a multilabel classification task. We use a maximum of 25 metadata tokens for the main entity span first inserting category, then description tokens. 

\paragraph{Relation Extraction} Relation extraction involves predicting the relation between the two substrings in the input example, one representing a subject and the other an object. We use the FewRel and TACRED (Revisited) benchmarks for evaluation \citep{han2018fewrel,zhang2017tacred, Alt2020TACREDRA}. FewRel consists of a set of 80 output relations and TACRED includes 42 possible output relations. We use a maximum of 20 metadata tokens per subject, object entity, first inserting category, then description tokens. 

For certain entity spans in these tasks, there are no linked entities (e.g., several are simply pronouns or dates). Table \ref{tab:statistics} gives task statistics. Section 4 considers different numbers and positions of inserted metadata tokens in the shaped examples.

\begin{table}[t!]
    \begin{center}
    
    \vspace{3mm}
    \begin{tabular}{llccc}
    \toprule
    Benchmark    &    Train  &   Valid &   Test  \\
    \midrule
    TACRED            &  68124      &   22631      &    15509  \\
    Category     &  54k/46k &  16k/15k & 9k/10k \\
    Description  &  50k/43k &  15k/14k & 8k/9k \\
    \midrule
    FewRel          & 8k & 16k &  16k \\
    Category    & 8k/8k  & 16k/15k & 16k/15k \\
    Description  & 7k/8k  & 15k/16k & 15k/16k \\
    \midrule
    OpenEntity  &  1998 & 1998 & 1998 \\
    Category &  674 & 674  & 647 \\
    Description & 655 & 672  & 649 \\
    \bottomrule
    \end{tabular}
    \caption{We show the benchmark split sizes (row 1), and the \# of examples tagged with category and description metadata (rows 2 and 3). We give numbers for the subject and object entity-span on relation extraction and the main entity-span for entity-typing. Tasks have different proportions of shaped examples.}
    \vspace{-0.4cm}
    \label{tab:statistics}
    \end{center}
\end{table}

\subsection{Experimental Settings}
\paragraph{Source of Metadata} We use a pretrained entity-linking model \citep{bootleg} to link the text in each task to an October 2020 dump of Wikidata. We use Wikidata and the first sentence of an entity's Wikipedia page to obtain descriptions. Additional details for the entity linking model and the metadata are in the appendix.

\paragraph{Model} We fine-tune a BERT-base model on metadata shaped data for each task, taking the pooled representation and using a linear prediction layer for classification \citep{devlin2018bert}. We use cross-entropy loss for FewRel and TACRED and binary-cross-entropy loss for
OpenEntity. All test scores are reported at the epoch with the best validation score and we use the scoring implementations released by \citep{zhang2019ernie}. Additional training details are in the appendix.

\subsection{Baselines}
The following knowledge-aware LMs are proposed by prior work for the evaluated tasks and currently state-of-the-art. \textbf{ERNIE}, \citep{zhang2019ernie} 
\textbf{LUKE} \citep{yamada2020luke},
\textbf{KEPLER} \citep{wang2020kepler},
\textbf{CokeBERT} \citep{su2021cokebert},
and \textbf{WKLM} \citep{xiong20wklm}
introduce auxilliary loss terms and require additional pretraining. 
Prior approaches also modify the architecture for example using  alternate attention mechanisms (\textbf{KnowBERT} \citep{peters2019knowledge}, \textbf{K-BERT}, \citep{liu2020kbert}, \textbf{LUKE}) or training additional transformer stacks to specialize in knowledge-based reasoning (\textbf{K-Adapter} \citep{wang2020kadapter}). \textbf{E-BERT} \citep{perner2020ebert} does not require additional pretraining and uses entity embeddings which are aligned to the word embedding space. 
In Table \ref{tab:benchmarks}, we compare to methods which use the same base LM, BERT-base, and external information resource, Wikipedia, for consistency.

\begin{table*}[t!]
\begin{center}
\begin{tabular}{l|lll|lll|lll}
    \toprule
                 \multirow{2}{*}{Model} & \multicolumn{3}{c|}{FewRel} & \multicolumn{3}{c|}{TACRED} & \multicolumn{3}{c}{OpenEntity}  
                 \\ \cline{2-10}
                 & P       & R       & F1      & P       & R       & F1      & P         & R        & F1 \\      
                 \cline{1-10}
BERT-base        & 85.1    & 85.1    & 84.9    & 66.3    & 78.7    & 72.0    & 76.4      & 71.0     & 73.2 \\    

\cline{1-10}

ERNIE            & 88.5    & 88.4    & 88.3    & 74.8    & 77.1    & 75.9    & 78.4      & 72.9     & 75.6  \\

E-BERT$_{concat}$  & 88.5    & 88.5    & 88.5    & -       & -       & -       & -         & -        & -     \\

K-BERT    & 83.1  & 85.9    & 84.3    & -       & -       & -       & 76.7     & 71.5    & 74.0   \\

KnowBERT$_{Wiki}$    & 89.2    & 89.2    & 89.2    & \textbf{78.9}      & \textbf{76.9}      & \textbf{77.9}       & 78.6      & 71.6     & 75.0   \\

CokeBERT    & 89.4    & 89.4    & 89.4    & -       & -       & -       & 78.8      & \textbf{73.3}     & 75.6    \\
\cline{1-10}

Ours (BERT-base) & \textbf{90.2}    & \textbf{90.2}     & \textbf{90.2}     & 75.9    & 75.6    & 76.3    & \textbf{79.3}      & \textbf{73.3}     & \textbf{76.2}     \\
    \bottomrule
\end{tabular}
\caption{Test scores on relation extraction and entity-typing tasks. ``Ours (Base LM)'' is metadata shaping. All methods use the same base LM (BERT-base) and external information (Wikipedia) for consistency.}
    \label{tab:benchmarks}
    \end{center}
\end{table*}

\subsection{End-to-End Benchmark Results}
\textbf{Main Results} We simply use an \textit{off-the-shelf} BERT-base LM \citep{wolf2020transformers}, with no additional pretraining and fine-tuned on shaped data. 
Directly using the same three metadata shaping functions for each task, metadata shaping exceeds the BERT-base LM trained on unshaped data by \yell{5.3} (FewRel), and \yell{4.3} (TACRED), \yell{3.0} (OpenEntity) 
F1 points. Metadata shaping is also competitive with SoTA baselines which \textit{do} modify the LM and use the same Wikipedia resource
(Table \ref{tab:benchmarks}). For the baselines, we use reported numbers when available and \citet{su2021cokebert} reports two KnowBERT-Wiki results. We obtain other numbers using the released code (See Appendix).

The metadata, which are at different granularities, are more useful in combination. Ablating each shaping function for FewRel and OpenEntity,  the task structure tokens, with no other shaping, provide a \yell{26.3} (FewRel), \yell{24.7} (OpenEntity) F1 point.
boost. Category tokens for the entities tagged in the examples, and no descriptions, provides a \yell{3.0} (FewRel), \yell{2.5} (OpenEntity) F1 point boost. 
Descriptions of entities tagged in the examples, with no category tokens, provides a \yell{1.6} (FewRel), \yell{1.5} (OpenEntity) F1 point boost.

%% file: Sections/principles.tex
Given the framework introduced in Section 2, here we study the following key questions for effectively using metadata shaping: \textbf{Framework (Section 4.1)} What are the reasons for different granularities of metadata? How should we select and combine metadata? 
\textbf{Tail Evaluation (Section 4.2)} What are the effects of metadata shaping on slices concerning tail vs. popular entities in the data?
\textbf{Insertion Choices (Section 4.3)} How do different choices for inserting metadata into original examples affect performance?

\subsection{Framework: Role of Metadata Types}
\paragraph{Metadata Effects} \textit{Class-discriminative metadata correlates with reduced model uncertainty. With high quality metadata, as found in Wikidata, this results in improved classification performance.}

First to investigate the effects of metadata on model uncertainty, we compute the entropy of $\hat{p}_{\phi}$ softmax scores over the output classes as a measure of uncertainty, and compute the average across test set examples. We observe that lower uncertainty by this metric is correlated with improved classification performance (See Figure \ref{fig:mlmcls} (Top)).

We compute pointwise mutual information ($\pmi$) scores for inserted metadata tokens as a measure of class-discriminativeness. We rank individual tokens $k$ by $\pmi(y, k)$ (for task classes $y$), computed over the training dataset. On FewRel, for test examples containing a top-20 $\pmi$ word for the gold class, the accuracy is \yell{27.6\%} higher vs. the slice with no top-20 $\pmi$ words for the class. Notably, \yell{74.1\%} more examples contain a top-20 $\pmi$ word for their class when $\pmi$ is computed on shaped data vs. when the training data is unshaped.

\paragraph{Metadata Token Selection} \textit{Simple information theoretic heuristics suffice to rank metadata by the information gain they provide, despite the complexity of the underlying contextual embeddings.}

We apply Algorithm \ref{CHalgorithm} to select metadata tokens for the tasks, which ranks metadata tokens by their provided information gain. Given an entity with a set $\pmb{M}$ of metadata tokens, our goal is to select $n$ to use for shaping. We compare four selection approaches: using the highest (``High Rank'') and lowest (``Low Rank'') ranked tokens by Algorithm \ref{CHalgorithm}, random metadata from $\pmb{M}$ (``Random''), and the most popular metadata tokens across the union of $\pmb{M(x_i)}, \forall x_i \in \pmb{D}$ (``Popular''), selecting the same number of metadata tokens per example for each baseline. We observe that ``High Rank'' consistently gives the best performance, evaluated over multiple seeds, and note that even ``Random'' yields decent performance vs. the BERT-baseline, indicating the simplicity of our method (Table \ref{tab:algorithm}).

\begin{table}[t!]
    \begin{center}
    \begin{tabular}{llccc}
    \toprule
    Benchmark   &  Strategy  &    Test F1  \\
    \midrule
    \multirow{5}{*}{FewRel}
    &     BERT-base     &        \yell{84.9}  \\
    &     Random     &           \yell{87.3 $\pm 0.8$}  \\
    &     Popular    &           \yell{87.9 $\pm 0.1$}  \\
    &     Low Rank   &           \yell{87.8 $\pm 0.4$}  \\
    &     High Rank  &   \textbf{\yell{89.0 $\pm 0.6$}}  \\
    \midrule
    \multirow{5}{*}{OpenEntity}
    &     BERT-base  &           \yell{73.2} \\
    &     Random     &           \yell{74.3 $\pm 0.7$}  \\
    &     Popular    &           \yell{74.5 $\pm 0.4$}  \\
    &     Low Rank   &           \yell{74.1 $\pm 0.4$}  \\
    &     High Rank  &   \textbf{\yell{74.8 $\pm 0.1$}}  \\
    \midrule
    \multirow{5}{*}{TACRED}
    &     BERT-base  &           \yell{72.0} \\
    &     Random     &           \yell{73.8 $\pm 1.6$}  \\
    &     Popular    &           \yell{73.6 $\pm 0.9$}  \\
    &     Low Rank   &           \yell{73.3 $\pm 1.0$}  \\
    &     High Rank  &   \textbf{\yell{74.7 $\pm 0.5$}}  \\
    \bottomrule
    \end{tabular}
    \caption{Average and standard deviation over 3 random seeds. Each method selects up to $n$ metadata tokens per entity. For FewRel, TACRED, $n = 3$ per subject, object. For OpenEntity $n = 2$ per main entity as \yell{33\%} of OpenEntity train examples have $\geq 2$ categories available (\yell{80.7\%} have $\geq 3$ categories on FewRel). Note we use larger $n$ for the main results (Table \ref{tab:benchmarks}).}
    \vspace{-0.4cm}
    \label{tab:algorithm}
    \end{center}
\end{table}

Considering the distribution of selected category tokens under each scheme, the KL-divergence between the categories selected by ``Low Rank'' vs. ``Popular'' is \yell{0.2} (FewRel), \yell{4.6} (OpenEntity), while the KL-divergence between ``High Rank'' vs. ``Popular'' is \yell{2.8} (FewRel), \yell{2.4} (OpenEntity). Popular tokens are not simply the best candidates; Algorithm \ref{CHalgorithm} selects discriminative  metadata. 

For OpenEntity, metadata are relatively sparse, so categories appear less frequently in general and it is reasonable that coarse grained types have more overlap with ``High Rank''. For e.g., ``business'' is in the top-10 most frequent types under ``High Rank'', while ``non-profit'' (occurs in 2 training examples) is in ``Low Rank'''s top-10 most frequent types. Metadata tokens overall occur more frequently in FewRel (See Table \ref{tab:statistics}), so finer grained types are also quite discriminative. The most frequent category under ``Low Rank'' is coarse grain, ``occupation'' (occurs in 2.4k training examples), but the top-10 categories under ``High Rank'' are finer grain, containing ``director'' and ``politician'' (each occurring in $>$ 300 training examples).

\paragraph{Metadata Noise} \textit{Noisier metadata can provide implicit regularization. Noise arises from varied word choice and order, as found in entity descriptions, or blank noising (i.e. random token deletion).} 

Descriptions use varied word choice and order vs. category metadata.\footnote{Over FewRel training data: on average a word in the set of descriptions appears \yell{8} times vs. \yell{18} times for words in the set of categories, and the description set contains \yell{3.3x} the number of unique words vs. set of categories.} 
To study whether shaping with description versus category tokens lead the model to rely more on metadata tokens, we consider two shaping schemes that us $10$ metadata tokens: 10 category tokens and 5 category, 5 description. We observe both give the $\sim$same score on FewRel, \yell{89.8} F1 and \yell{89.5} F1, and use models trained with these two schemes to evaluate on test data where $10\%$ of metadata tokens per example are randomly removed. Performance drops by \yell{1.4} F1 for the former and \yell{1.0} F1 for the latter. 

Blank noising \citep{xie2017blanknoising} by randomly masking 10\% of inserted metadata tokens during training leads to a consistent boost on OpenEntity: \yell{0.1} (``High Rank''), \yell{0.5} (``Popular''), \yell{0.5} (``Low Rank'') F1 points higher than the respective scores from Table \ref{tab:algorithm} over the same 3 random seeds. We observe no consistent benefit from masking on FewRel. Future work could investigate advanced masking strategies, for example masking discriminative words in the training data.

\paragraph{Task Agnostic Metadata Effects} \textit{Using metadata correlates with reduced task-specific model uncertainty, discussed in the prior analysis of the shaped model's softmax scores. Interestingly, metadata also correlates with reduced LM uncertainty in a task-agnostic way.}

Metadata shaping requires no additional pretraining for downstream tasks. However, in this section alone, we consider the effects of additional pretraining using shaped data on the base LM. Here, we perform additional masked language modeling (MLM) over the shaped task training data for the off-the-shelf BERT model to learn model $\hat{p}_{\theta}$. We minimize the following loss function and evaluate the model perplexity on the task test data:

\begin{figure}[t!]
    \centering
    \includegraphics[width=0.45\linewidth]{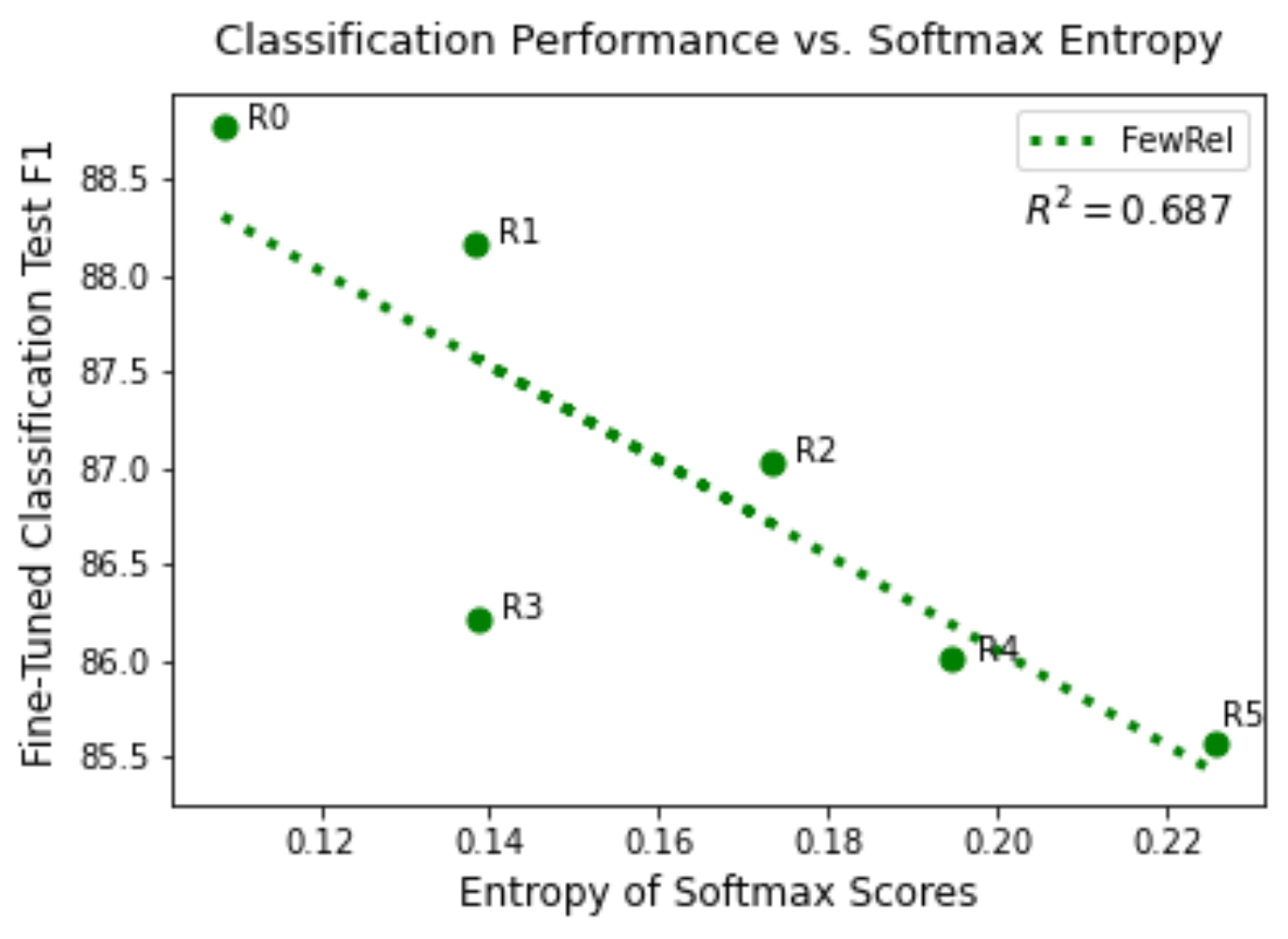}
    \includegraphics[width=0.45\linewidth]{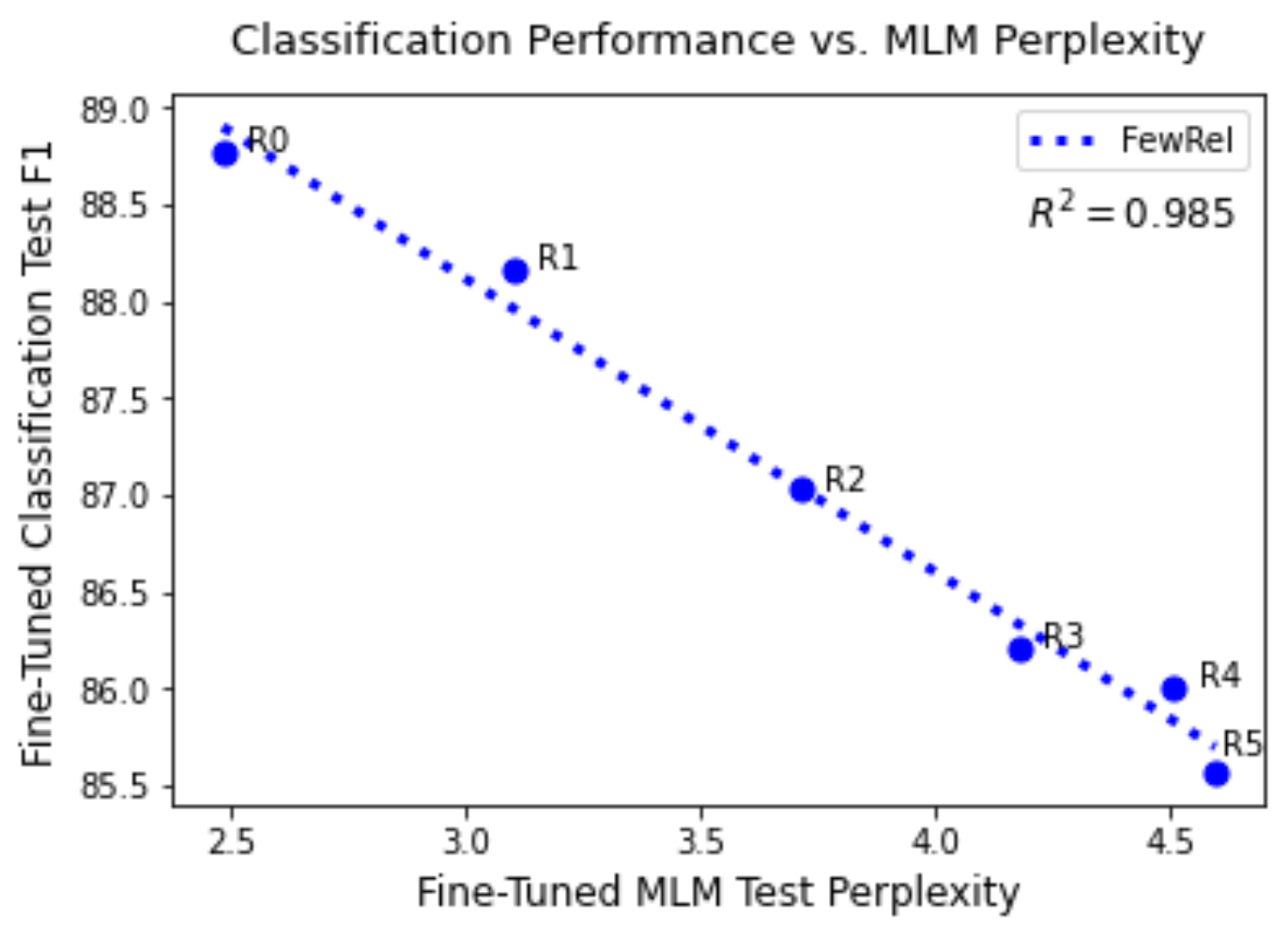}
    \caption[width=0.9\linewidth]{Test F1 for $\hat{p}_{\phi}$ (no additional pretraining) vs. average entropy of $\hat{p}_{\phi}$ softmax scores (Top) and vs.  perplexity of an independent model $\hat{p}_{\theta}$ (w/ additional pretraining) (Bottom). $\hat{p}_{\phi}$ and $\hat{p}_{\theta}$ use the same shaped training data. Each point is a different metadata shaping scheme (median over 3 random seeds): for \emph{R0} all inserted tokens are true category tokens associated with the entity. For \emph{RX}, X true metadata tokens are replaced by randomly chosen tokens from the full vocabulary (i.e., noise, unlikely to be class-discriminative). For each point, the total number of metadata tokens (true plus random) is held constant per example.}
    \label{fig:mlmcls}
\end{figure}

\begin{table}[t!]
    \begin{center}
    \begin{tabular}{llccc}
    \toprule
    Benchmark   &  $R^2$  \\
    \midrule
    FewRel     &   \yell{0.985}  \\
    TACRED    &     \yell{0.782*}  \\
    OpenEntity  &   \yell{0.956}  \\
    \bottomrule
    \end{tabular}
    \caption{Correlation ($R^2$) between test F1 of $\hat{p}_{\phi}$ (no additional pretraining) vs. perplexity of $\hat{p}_{\theta}$ (w/ additional pretraining) for three tasks, using the procedure described in Figure \ref{fig:mlmcls}. *Without one outlier corresponding to shaping with all random tokens ($R^2 = 0.02$ with this
    point).}
    \vspace{-0.4cm}
    \label{tab:r2perplexity}
    \end{center}
\end{table}

{\small\begin{align}
    \mathcal{L}_{\textup{mlm}} = \mathbb{E}_{s \sim D, m\sim M, i\sim I} \big[ -\log(\hat{p}_{\theta}(s_{m_i}|s_m/i))\big].
\end{align}}

where $I$ is the masked token distribution and $s_{m_i}$ is the masked token at position $i$ in the shaped sequence $s_m$.\footnote{We use the Hugging Face implementation for masking and fine-tuning the BERT-base MLM \citep{wolf2020transformers}.} Through minimizing the MLM loss, $\hat{p}_\theta$ learns direct dependencies between tokens in the data \citep{zhang2021mlmdeps}. In Figure \ref{fig:mlmcls} (Bottom), Table \ref{tab:r2perplexity}, we observe a correlation between reduced perplexity for $\hat{p}_{\theta}$, and higher downstream performance for $\hat{p}_{\phi}$ across multiple tasks, both using the same training data. Lower perplexity indicates a higher likelihood of the data. It is interesting to observe this correlation since standard language modeling metrics such as perplexity differ from downstream metrics.

\begin{figure}[t!]
    \centering
    \includegraphics[width=0.6\linewidth]{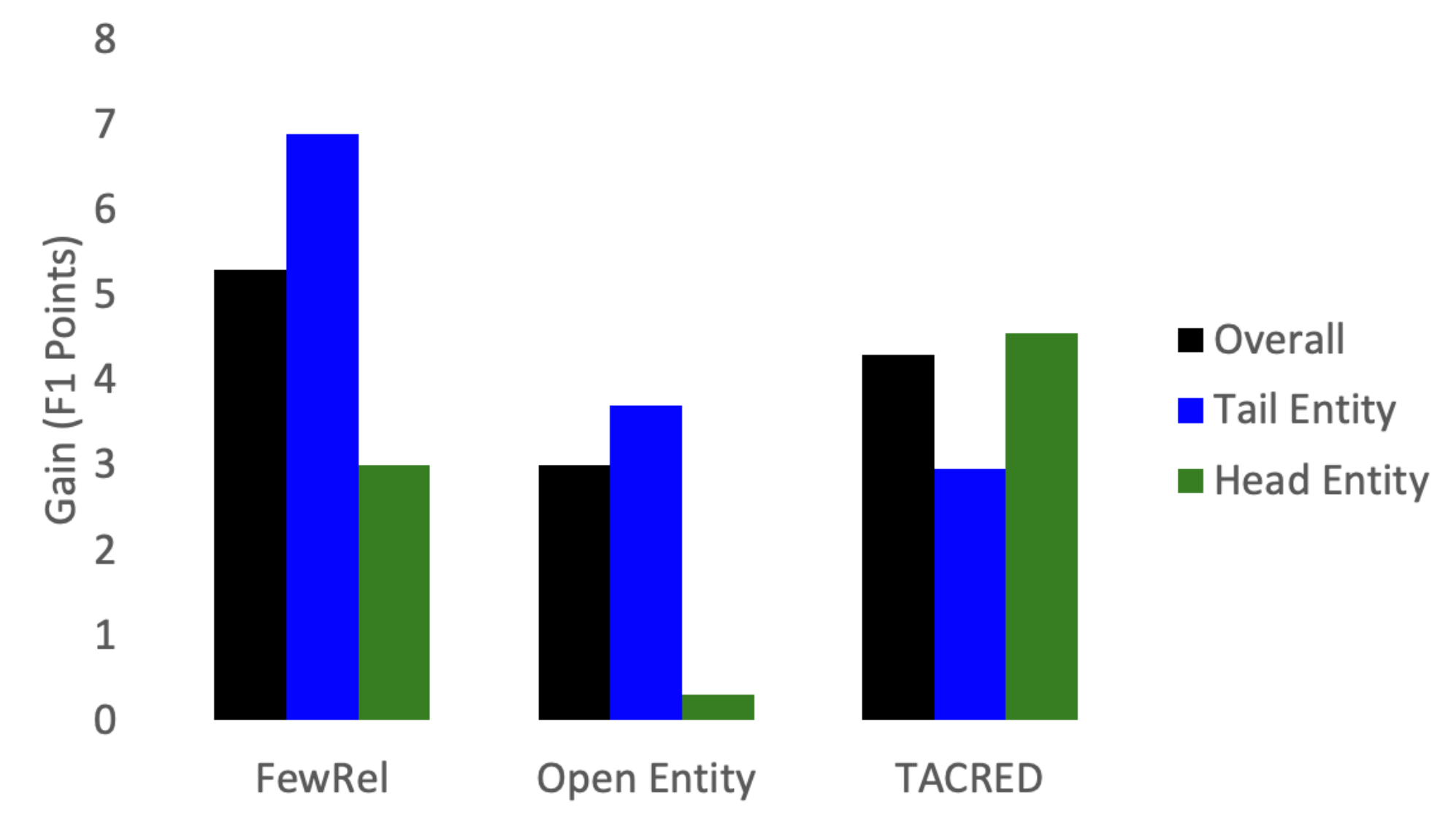}
    \caption[width=0.9\linewidth]{The gain from the LM trained with metadata shaping vs. the BERT-base baseline split by the popularity of the entity span in the example.}
    \label{fig:popularity_slice}
\end{figure}

\subsection{Evaluation: Tail and Head Slices}
Section 3 shows overall improvements from shaping.  Here we consider fine-grained slices: the ``tail'' slice contains entity spans seen $<10$ times and ``head'' slice contains those seen $\geq 10$ times during training. We observe that much of the gain from shaping occurs on the tail test data (Figure \ref{fig:popularity_slice}). 

For the unshaped BERT-base LM, the test F1 score is \yell{93.3} (FewRel), \yell{80.7} (OpenEntity) points on head examples, but just \yell{82.5} (FewRel), \yell{69.6} (OpenEntity) points on the tail. The shaped LM improves the tail F1 score by \yell{6.9} (FewRel), \yell{3.0} (OpenEntity) points and head F1 score by \yell{3.7} (FewRel), \yell{0.3} (OpenEntity) points. For TACRED, the base test F1 score is \yell{68.7} on head examples and \yell{73.7}  on the tail. The shaped LM improves the tail F1 score by \yell{3.0} and head F1 score by \yell{4.6}. A consideration for TACRED is that \yell{42\%} of these head spans are stopwords (e.g., pronouns) or numbers; just \yell{7\%} are for FewRel. \footnote{Based on unseen object spans for FewRel and TACRED, as $>$ 90\% subject spans are unseen.}

\paragraph{Subpopulations} \textit{Metadata are helpful on the tail as they establish subpopulations.}

For the tail, we hypothesize that metadata establish subpopulations. Thus, if a pattern is learned for an entity-subpopulation occurring in the training data, the model may perform better on rare entities that also participate in the subpopulation, but were not individually observed during training. On FewRel, we take the top-20 TF-IDF words associated with each category signal during training as linguistic cues captured by the model for the category subpopulation. 
For example, ``government'' is in the top-20 TF-IDF words associated with the ``politician'' entity category. At test time, we select a slice of examples containing any of these words for any of the categories inserted in the example. Using  metadata, the performance is \yell{9.0}, \yell{3.5} F1 points higher on examples with unseen subject, object entities with vs. without a top-20 TF-IDF word for a subject, object category.

\paragraph{Metadata Effects on Popular Entities} \textit{For popular entities the LM can learn entity-specific patterns well, and be mislead by subpopulation-level patterns corresponding to metadata.} 

Although we observe overall improvement on the popular entity slice, here we study the effect of metadata on popular entities within our conceptual framework. Let $p$ be a popular pattern (i.e., an entity alias) in the training data, and let $m$ be a metadata token associated with $p$.  Intuitively, the LM can learn entity-specific patterns from occurrences of $p$, but coarser grained subpopulation-level patterns corresponding to $m$. If $m$ and $p$ are class-discriminative for different sets of classes, then $m$ can mislead the LM.

To evaluate this, consider subject and object entity spans $p \in P$ seen $\geq 1$ time during training. For test examples let $\mathcal{Y}_{p}$ be the set of classes $y$ for which there is a $p \in P$ in the example with positive $\pmi(y, p)$, and define $\mathcal{Y}_{m}$ as the classes $y$ for which there is a metadata token $m$ with positive $\pmi(y, m)$ in the example. The set of examples where $\mathcal{Y}_p \neq \emptyset$. $\mathcal{Y}_m \neq \emptyset$, and $\mathcal{Y}_{p}$ contains the true class, but $\mathcal{Y}_{m}$ does not, represents the slice where metadata can mislead the model. On this slice of FewRel, the improvement from the shaped model is \yell{2.3 F1 points} less than the improvement on the slice of all examples with $\mathcal{Y}_p \neq \emptyset$. $\mathcal{Y}_m \neq \emptyset$. When we narrow to $p$ seen $\geq 10$ times during training, the error rate from the shaped model is actually \yell{1.1 F1 points} \textit{larger} than the unshaped model on the slice where $\mathcal{Y}_{p}$ contains the true class, but $\mathcal{Y}_{m}$ does not, supporting our intuition. 

An example of entity-specific vs. subpopulation-level patterns in FewRel is: $p=$ ``Thames River'' is class-discriminative for $y=$``located in or next to body of water'', but its $m=$``river'' is class-discriminative for $y=$``mouth of the watercourse''.

\subsection{Implementation Choices}
This section studies the sensitivity of metadata shaping to the location in the example where metadata are inserted (e.g., the impact of position, long sequence lengths, and special tokens).

\paragraph{Token Insertion} \textit{We observe low sensitivity to increasing the context length and to token placement (i.e., inserting metadata directly-following the entity-span vs at the end of the sentence).}

We evaluate performance a the maximum number of inserted tokens per entity, $n$, increases. On FewRel, OpenEntity, performance tends to improve with $n$.\footnote{Per subject and object entity for FewRel, and per main entity for OpenEntity. I.e., $n=10$ for FewRel yields a maximum of 20 total inserted tokens for the example.} For FewRel, $n \in$ \{1, 5, 10, 20, 25\} gives $\{85.4, 86.4, 87.6, 88.5, 90.2\}$ test F1. On OpenEntity, $n \in$ \{1, 5, 10, 20, 25, 40\} gives $\{74.9, 75.7, 74.8, 74.5, 76.2, 75.8\}$ test F1. Shaped examples fit within the max sequence length and we observe low sensitivity to longer contexts.

The benefit of inserting metadata directly-following the entity span vs at the end of the example differed across tasks (e.g., for TACRED, placement at the end performs better, for the other tasks, placement directly-following performs better), though the observed difference is $<1$ F1 point. We use the better-performing setting in Section 3 and for results on all tasks in Section 4, tokens are inserted directly-following the relevant entity span.

\paragraph{Boundary Tokens} \textit{Designating the boundary between original tokens in the example and inserted metadata tokens improves model performance.}

Inserting boundary tokens (e.g., ``\textbf{\#}'') in the example, at the start and end of a span of inserted metadata, consistently provides a boost across the tasks. Comparing performance with metadata and boundary tokens to performance with metadata and no boundary tokens, we observe a \yell{0.7 F1} (FewRel), \yell{1.4 F1} (OpenEntity) boost in our main results. We use boundary tokens for all results in this work.


%% file: Sections/discussion.tex
We ultimately find that in certain cases, we can match the quality of knowledge-aware architectures using a base LM, and only changing the data.  Benefits from metadata shaping include the ability to use simpler LMs, tackle the tail with better data efficiency by learning subpopulation-level (rather than only entity-specific) patterns, and provide a conceptual understanding by relying on the rich set of tools for reasoning about data. KnowBERT \citep{peters2019knowledge} considers NER, Wikpedia, and Wordnet \citep{miller1995wordnet}, and \citet{Alt2020TACREDRA} reports improvements from using NER and POS metadata, but neither explains the benefit or selection process.

However, there are also limitations to only changing the data. For instance, pretraining entity embeddings on a large number of examples allows the embedding to encode facts beyond the scope of the provided metadata tokens, and metadata shaping and the baselines (e.g., \citet{zhang2019ernie}) rely on access to accurate sources of external knowledge such as the entity-linking model to obtain entity information.
Finally, this work leverages bidirectional base LMs and focuses on classification --- in future work we want to study how to best incorporate metadata with different types of LMs and tasks.

%% file: Sections/related_work.tex
 \paragraph{Incorporating Knowledge in LMs}
As previously discussed, significant prior work incorporates knowledge into language modeling through changing the LM architecture or loss \citep{logan2019kglm, zhang2019ernie, peters2019knowledge, liu2020kbert, wang2020kadapter, yamada2020luke, xiong20wklm, wang2020kepler, su2021cokebert}, in contrast to our data focused approach. 
 
 \paragraph{Feature Selection} This work is inspired by  techniques in feature selection based on information gain \citep{ guyon2003ftrselection}. In contrast to traditional NLP feature schema \citep{levin1993verbannotations, marcus1993treebank}, metadata shaping annotations are expressed in natural language and the framework incorporates structured and unstructured annotations. We show classic feature selection methods based on information gain \citep{berger1996maxent} are informative even when using complicated contextual embeddings. Information gain is a standard feature selection principle, and in the setting we study --- the tail of entity-rich tasks --- we contribute an application of the principle to explain how metadata can reduce generalization error.

\paragraph{Prompt-based Learning} 
Prompting typically involves changing the downstream task template to better elicit \textit{implicit} knowledge from the base LM  \citep{liu2021promptsurvey}. In contrast, the same metadata shaping functions apply directly to different task types and the method focuses on how to \textit{explicitly} incorporate new, retrieved signals, not found in the original task. Prior work uses structured tokens in prompts for controllable text generation \citet{keskar2019ctrl, aghajanyan2021htlm}; we consider a broader variety of structured data and provide intuition for selecting between structured data.

\paragraph{Data Augmentation} One approach to tackle the tail is to generate additional examples for tail entities \citep{wei2019eda, xie2020uda, da2020neraug}. However, this can be sample inefficient \citep{horn2017tail} since augmentations do not explicitly signal that different entities are in the same subpopulation, so the model would need view each entity individually in different contexts. In contrast, shaping does not generate new samples. Metadata shaping and prompting \citep{scao2021dataaugprompt} may be viewed as implicit data augmentation.

\paragraph{Structured Data for Tail Generalization} 
Prior work leverages metadata for entities. Particularly relevant, \citet{Joshi2020tek, logeswaran2019entdesc} demonstrate improvements on QA and named entity disambiguation (NED) respectively by using descriptions of entities, and \citet{raiman2018deeptype} uses entity types for entity-matching. \citet{bootleg} demonstrates that category metadata improves tail performance for NED, but requires custom architectures and storing trained embeddings for all categories and entities. We combine different granularities of metadata, provide a method that applies generally to downstream classification tasks, do not modify the LM, and provide conceptual grounding.

%% file: Sections/conclusion.tex
 We propose metadata shaping to improve tail performance. The method is simple and general, yet competitive to model-based approaches for entity-rich tasks. We further show the empirical benefits are significant on the tail and formalize why metadata can reduce generalization error. While this work focused on entity-rich tasks, the method is not limited to this setting and we hope this work motivates further research on addressing the tail challenge through the data.

%% file: Sections/appendix.tex
\section{Appendix}
\subsection{Dataset Details}

\paragraph{Metadata} We tag original dataset examples with a pretrained entity-linking model from \citep{bootleg},\footnote{https://github.com/HazyResearch/bootleg} which was trained on an October 2020 Wikipedia dump with train, validation, test splits of 51M, 4.9M, and 4.9M sentences. FewRel includes entity annotations. The types we use as category metadata for all tasks are those appearing at least 100 times in Wikidata for entities their Wikipedia training data. This results in 23,413 total types. Descriptions are sourced from Wikidata descriptions and the first 50 words of the entity Wikipedia page. 

\paragraph{Benchmarks} Benchmark datasets are linked here: https://github.com/thunlp/ERNIE. 

\subsection{Training Details}
We use the pretrained BERT-base-uncased model for each task to encode the input text and a linear classification layer, using the pooled output representation. For TACRED, we also use a pretrained SpanBERT-large-cased model \citep{spanbert}.

\paragraph{Entity Typing} Hyperparameters include seed $1$, $2e-5$ learning rate, no regularization parameter and $256$ maximum sequence length. 
For OpenEntity we use a batch size of 16 and use no gradient accumulation or warmup. 

\paragraph{Relation Extraction}  Hyperparameters include seed $42$, $2e-5$ learning rate, no regularization parameter. For FewRel, we use  For FewRel, we use batch size 16, 512 max sequence length, and no gradient accumulation or warmup. For TACRED, we use 256 maximum sequence length, batch size 48, and no gradient accumulation or warmup. 
We report the test score for the epoch with the best validation score for all tasks.

For Section 4 experiments, the random seeds for tasks are $\{42, 4242, 1010\}$ for all tasks.

\subsection{Baseline Implementations}

We produce numbers for key baselines which do not report for our benchmarks, using provided code.\footnote{https://github.com/allenai/kb} \footnote{https://github.com/thunlp/ERNIE}
\begin{itemize}
    \item We produce numbers for KnowBERT-Wiki on TACRED-Revisited using a learning rate of $3e-5$, $\beta_2 = 0.98$, and choosing the best score for epochs $\in$ {1, \textbf{2}, 3, 4} and the remaining provided configurations used for the original TACRED task.
    \item We produce numbers for ERNIE on TACRED-Revisited using the provided training script and configurations they use for the original TACRED task.
\end{itemize}


